%% file: main.tex
\documentclass[10pt]{article}
\usepackage{cmap}
\input{hicss-packages.tex}
\usepackage[T1]{fontenc}
\usepackage{xurl}

\appto{\bibsetup}{\raggedright}

\makeatletter
\renewcommand\subsubsection{\@startsection{subsubsection}{3}{\z@}%
  {1.5ex plus 1ex minus .2ex}{-1em}%
  {\normalfont\normalsize\bfseries}}
\makeatother

\title{A Decision-Support Audit Protocol for Supervision Drift in
Proxy-Labeled Credit-Risk Prediction}

\author{%
\textbf{Mehrdad Shoeibi}\textsuperscript{1\,\authororcid{0009-0006-5470-0397}}\quad
\textbf{Muhammad Shabanpour}\textsuperscript{2\,\authororcid{0009-0005-4737-5808}}\quad
\textbf{Waldemar Karwowski}\textsuperscript{1\,\authororcid{0000-0002-9134-3441}}\quad
\textbf{Niloofar Yousefi}\textsuperscript{1\,\authororcid{0009-0005-6018-0634}}\\[8pt]
\textsuperscript{1}Department of Industrial Engineering and Management Systems\\
University of Central Florida, Orlando, FL, USA\\[4pt]
\textsuperscript{2}Department of Economics, Northeastern University, Boston, MA, USA\\[8pt]
\underline{Mehrdad.Shoeibi@UCF.edu}\quad \underline{shabanpour.m@northeastern.edu}\\[2pt]
\underline{wkar@ucf.edu}\quad \underline{niloofar.yousefi@ucf.edu}%
}

\date{}

\usepackage{microtype}
\usepackage{dblfloatfix}

\usepackage{flushend}

\usepackage[hidelinks]{hyperref}
\usepackage{orcidlink}
\newcommand{\authororcid}[1]{\orcidlink{#1}}
\hypersetup{
  pdftitle={A Decision-Support Audit Protocol for Supervision Drift in Proxy-Labeled Credit-Risk Prediction},
  pdfauthor={Mehrdad Shoeibi; Muhammad Shabanpour; Waldemar Karwowski; Niloofar Yousefi},
  pdfsubject={Proceedings of the 60th Hawaii International Conference on System Sciences (HICSS-60)},
  pdfkeywords={supervision drift; credit-risk prediction; proxy labels; model calibration; design science},
}

\begin{document}
\maketitle

\begin{abstract}
Credit-risk models are trained on proxy labels and deployed under temporal and segment
change, yet no single transfer metric separates base-rate shift, probability-scale shift, and
feature-label relationship change. We contribute a design-science artifact: a locked,
multi-signal audit protocol for supervision drift in proxy-labeled credit-risk prediction.
Five layers (transfer performance, an oracle-gap probe, a calibration diagnostic,
feature-label stability, and a synthetic positive control), thresholds, and decision rules
were locked before interpretation; a bounded reading is a designed outcome. On a public
LendingClub dataset (temporal 2013 to 2016 and cross-segment transfer), ranking is stable and
oracle gaps are small; the clearest temporal signal is a prevalence and probability-scale
mismatch that intercept-only diagnostic recalibration largely reduces, though its cause is not
identifiable from the available release. The positive control responds only to larger injected
shifts; subtler drift cannot be excluded. Mapping diagnostic patterns to governance actions is
conceptual guidance, not validated here.
\end{abstract}

\subsubsection*{Keywords:}

Supervision drift, credit-risk prediction, proxy labels, model calibration, design science.

\section{Introduction}

Machine learning systems for credit-risk prediction are trained on historical lending records and then
applied to applicants from later periods or different market segments. The outcome label that supervises
these systems, such as whether a granted loan defaulted, is an operational proxy for realized credit
risk rather than a direct measure of creditworthiness. Two features complicate evaluation: the
supervision signal is a proxy whose reliability may change over time, and the deployment environment is
non-stationary as macroeconomic conditions, lending policies, and borrower behavior evolve. Production
analyses have long identified changes in the external world as a primary and often hidden source of risk
\autocite{sculley2015hidden}, and the literature on dataset shift and concept drift documents how a model
validated on one distribution can behave differently on another
\autocite{quinonero2009dataset,gama2014survey,koh2021wilds}.

We use the term supervision drift, following its formalization as a change in $P(y \mid x, c)$ across
contexts \autocite{shoeibi2026learning}, for relationship-relevant change in this proxy-labeled setting:
change that affects transfer performance, the recoverability of performance through target-context
training, calibration, marginal feature-label association, or sensitivity to injected shifts. This is
narrower than the full taxonomy of distribution shift \autocite{morenotorres2012unifying}; pure
feature-distribution change matters only insofar as it affects these layers.

The stakes are not only financial. Credit scores govern access to housing, education, and working
capital, and a model whose probability scale has drifted can misprice risk even while its ranking appears
intact, producing unequal terms or denials that are difficult to detect from aggregate accuracy alone.
Misdiagnosis carries its own cost: a model left in place under a mistaken reading continues to price
applicants on a scale that no longer holds.

A central difficulty is that a single transfer metric cannot characterize supervision drift.
Discrimination and calibration are distinct, so a model can rank applicants in nearly the same order
while the mapping from features to realized default probabilities shifts. Reporting only a ranking
metric such as AUROC can understate a real probability-scale change, whereas reporting only a proper
scoring rule such as the Brier score can overstate relationship change, since it also absorbs base-rate
and scale differences. Distinguishing these explanations requires several complementary signals
interpreted together.

This paper makes a design-science contribution: a locked, layered, multi-signal diagnostic protocol for
auditing supervision drift in proxy-labeled credit-risk prediction. The protocol combines five
complementary layers: transfer performance, an oracle-gap probe of target-learnable change, a
calibration diagnostic, feature-label stability, and a synthetic positive control. The transfer settings, the model and metric plan, the numeric thresholds, and the
A through G decision rules were locked before the outputs were interpreted, which limits post-hoc
storytelling and makes a bounded reading an intended outcome rather than a failure. Section~3.1 states the design principles that organize the artifact.

The intended setting is decision support for model-risk governance: analysts and credit-risk managers
who must decide what to do with a deployed scoring model as conditions change. The risk the protocol
addresses is treating a stable ranking metric as sufficient evidence of safe deployment when the
probability scale on which credit decisions and pricing depend may have shifted.

We evaluate the protocol on a single large public LendingClub dataset, across a temporal transfer
setting (train on the 2013 vintage, evaluate on the 2016 vintage) and a cross-segment setting (train on
the 2013 debt-consolidation segment, evaluate on the 2013 credit-card segment). The empirical scope is
intentionally one dataset, which preserves analytical depth but limits external generalization. The
diagnostics indicate stable ranking transfer and small oracle gaps, with the clearest signal a temporal
prevalence and probability-scale mismatch that a diagnostic recalibration largely reduces. We report
this as bounded evidence rather than as proof that the conditional relationship is unchanged, and we do
not attribute the mismatch to a specific mechanism.

\section{Related Work}

Our study draws on several lines of research: the characterization of dataset shift and
concept drift, predictive modeling and shift handling in credit scoring, probability
calibration, and weak or proxy supervision. We review each in turn and then position our
contribution, which is an audit protocol rather than a new model or adaptation method.

\subsection{Dataset shift and concept drift}

The conditions under which a model trained on one distribution fails on another are well studied.
\textcite{quinonero2009dataset} gave a foundational treatment of dataset shift,
\textcite{morenotorres2012unifying} a unifying taxonomy distinguishing covariate, prior-probability, and
concept shift, and \textcite{gama2014survey} a survey of concept drift and adaptation, mainly in
streaming settings. These establish the vocabulary this study relies on. Rather than proposing a new detector or
adaptation algorithm, we fix a proxy-labeled credit-risk task and use complementary layers to separate
ranking transfer, probability-scale mismatch, marginal feature-label change, and target-learnable
multivariate change. \textcite{koh2021wilds} introduced WILDS and showed standard training yields
substantially lower out-of-distribution performance; we instead characterize the signature of any
observed gap on tabular credit-risk data rather than building a more robust predictor.

\subsection{Credit scoring and shift in credit scoring}

Within credit scoring, model comparison has a long history:
\textcite{lessmann2015benchmarking} benchmarked forty-one classifiers across eight credit-scoring
datasets, and \textcite{schmitt2022deep} found gradient boosting generally preferable to deep learning
on structured credit data. These treat the classifier as the object of interest; we instead use logistic
regression, histogram gradient boosting, and random forests as diagnostic instruments, not to identify a
best model.

A smaller body of work addresses distribution shift in credit scoring directly.
\textcite{qian2022managing} used adversarial validation to match training and target distributions on
LendingClub data, noting leakage risk when time periods are mixed, and \textcite{rahman2026modeling}
examine deployment reliability under temporal shift in LendingClub-style settings. Our focus is
complementary: rather than prescribing retraining or recalibration policies, we audit what type of shift
the evidence supports under a fixed proxy-label construction. Reject inference is a related concern;
\textcite{kozodoi2019shallow} proposed a self-learning framework to reduce sample-selection bias, whereas
we hold the proxy-label construction fixed rather than inferring missing labels.

\subsection{Calibration and label shift}

Probability reliability is a distinct evaluation dimension from ranking quality.
\textcite{platt1999probabilistic} introduced a parametric mapping from classifier scores to calibrated
probabilities, \textcite{niculescumizil2005predicting} showed that learning algorithms produce
systematically distorted probabilities correctable post hoc, and \textcite{guo2017calibration}
demonstrated that accurate modern classifiers can still be poorly calibrated, introducing temperature
scaling. Closely connected is the literature on label, or prior probability, shift:
\textcite{lipton2018detecting} proposed Black Box Shift Estimation to detect and correct changes in the
label marginal while the class-conditional distribution is assumed unchanged. In our protocol, calibration is a diagnostic layer identifying probability-scale and prevalence
mismatch under transfer, applied in an intentionally oracle-style form because it uses the observed
target default rate for analysis rather than production.
Whereas label-shift correction assumes the shift is of that type and corrects it, our protocol treats
prevalence shift as one candidate mechanism among several and tests whether the observed temporal
miscalibration is consistent with it.

\subsection{Weak and proxy supervision}

The reliability of the supervision signal is central here, since our outcome label is a proxy for credit
risk rather than observed ground truth. \textcite{ratner2016data} introduced data programming for
constructing training sets from heuristic labeling functions, implemented in the Snorkel system
\autocite{ratner2017snorkel}. That line concerns label construction; our study is downstream, asking
whether a fixed proxy supervision remains reliable under temporal and segment shift. Recent work
formalizes supervision drift as a change in $P(y \mid x, c)$ across contexts and studies it in a
weak-supervision genomics setting \autocite{shoeibi2026learning}. We adopt this conceptual framing but
evaluate a different artifact: a locked diagnostic audit protocol for proxy-labeled credit-risk
prediction.

\subsection{Decision support, model-risk governance, and design science}

The audit is framed as a decision-support artifact, connecting it to two information-systems literatures.
The first is model-risk governance and the monitoring, validation, and recalibration of deployed scoring
models, where the question is not only accuracy but what action a change in model behavior should
trigger. The second is design-science research, which treats a purposeful, evaluated artifact as
the unit of contribution and provides the rigor-and-relevance framing, methodology, and
contribution-positioning we adopt \parencite{hevner2004design,peffers2007design,gregor2013positioning}.
This positioning makes the bounded result legible as an artifact behaving correctly rather than a weak
empirical finding.

\subsection{Positioning}

Taken together, prior work motivates evaluating credit-risk models under shift but leaves open how to
interpret evidence when the labels are proxy-based and the target construction is fixed. We address this
gap with a pre-specified, multi-signal audit protocol designed to separate stable ranking transfer from
probability-scale mismatch and from stronger evidence of relationship-relevant drift.

\section{Methods}

\subsection{The artifact as a design-science contribution}

This work follows the design-science research paradigm, in which the primary contribution is a
purposeful artifact and a demonstration of its behavior, evaluated for rigor and relevance
\parencite{hevner2004design,peffers2007design}. Following \textcite{gregor2013positioning}, we position the contribution as an improvement: a new way to solve a recognized problem, interpreting deployment-time
movement in proxy-labeled credit-risk models, for which single-metric practice is inadequate. The
artifact is the locked, multi-signal audit protocol; the LendingClub study is an instantiation
demonstrating how it behaves on a real proxy-labeled task, not a generalization about credit markets.

The artifact is governed by four design principles. \emph{Multi-signal disconfirmation}: because no
single transfer metric can separate base-rate shift, probability-scale shift, and relationship change,
the protocol requires convergent evidence and treats a conclusion as supported only when the layers
agree. \emph{Pre-commitment}: all settings, metrics, thresholds, and rules are locked before any output
is interpreted, so a bounded reading is a designed outcome. \emph{Mechanism separation}: a base-rate or
probability-scale shift, which recalibration can address, is made distinguishable from relationship
change, which would call for retraining. \emph{Explicit sensitivity boundary}: a synthetic positive
control characterizes the magnitude of injected shift the stack could have detected, so absence of
large-drift evidence is reported with a stated detection floor rather than asserted as absence of drift.

Evaluation is by demonstration and analysis: the protocol is applied to a real proxy-labeled task across
temporal and cross-segment transfer, and the positive control provides an internal sensitivity analysis.
The evaluation target is the diagnostic protocol, namely whether the layers characterize the shift
signature they are designed to separate. It is not a test of whether the associated governance actions
are the correct organizational choices, which would depend on materiality and cost the protocol does not
model; that mapping is conceptual guidance (Section~5.3), not a validated decision rule.

\subsection{Empirical setting and prediction target}

We study supervision drift in a proxy-labeled tabular credit-risk setting using one public LendingClub
dataset \autocite{arizagarzon2024lendingclub}. The prediction target is a binary operational proxy for
realized credit risk. The proxy-label construction is fixed throughout, so the analysis concerns whether
the relationship between features, proxy labels, and predictive behavior remains stable enough for
transfer.

The protocol focuses on supervision drift and relationship-relevant shift, meaning changes that affect
the five layers of Section~3.5. Pure covariate shift in $P(x)$ is not a standalone diagnostic target; it
enters only insofar as it affects those layers.

\subsection{Transfer settings}

Each evaluation compares a source context, on which a model is trained, with a target context, on which
it is evaluated. The protocol uses an in-domain baseline plus two transfer settings, all locked before
the outputs were interpreted. The in-domain baseline trains and evaluates within the same context using a
held-out fold. The temporal setting uses the 2013 loan vintage as source and the 2016 vintage as target;
the cross-segment setting uses the 2013 debt-consolidation segment as source and the 2013 credit-card
segment as target. Table~\ref{tab:settings} reports the sizes and observed proxy-label prevalences; the
settings differ in prevalence and sample size, which is relevant to the interpretation boundaries below.

\begin{table*}[!tb]
\centering
\caption{Locked evaluation settings. Prevalence is the observed proxy-label (\texttt{Default}) rate.}
\label{tab:settings}
\begin{tabular}{l l r r r r}
\hline
\textbf{Context} & \textbf{Source $\rightarrow$ Target} & \textbf{Train n} & \textbf{Test n} & \textbf{Train prev.} & \textbf{Test prev.} \\
\hline
in\_domain & 2013 training fold $\rightarrow$ 2013 held-out fold & 107{,}843 & 26{,}961 & 0.1560 & 0.1560 \\
temporal & 2013 vintage $\rightarrow$ 2016 vintage & 134{,}804 & 293{,}057 & 0.1560 & 0.2328 \\
cross\_segment & 2013 debt consolidation $\rightarrow$ 2013 credit card & 80{,}634 & 32{,}804 & 0.1636 & 0.1320 \\
\hline
\end{tabular}
\end{table*}

\subsection{Models and evaluation metrics}

To avoid relying on a single learner, we use three model families: logistic regression, random forest,
and histogram gradient boosting, spanning a linear model and two nonlinear tree-based models without
exhausting all classes. Models are trained on the source context and evaluated on the target context,
with random forest run across several seeds and the other learners under a fixed seed, as specified in
the locked protocol.

Predictive behavior is summarized with AUROC for ranking, average precision under class imbalance, and
the Brier score for probability-scale error. Calibration is characterized using expected calibration
error (ECE) under equal-frequency bins, plus a calibration intercept and slope from a logistic regression
of the outcome on the model logit after clipping.

\subsection{Layered diagnostic protocol}

The protocol is a layered design rather than a list of unrelated metrics. Each layer answers a distinct
question about transfer behavior and carries an explicit boundary on what it can support. The layers feed
the rule-based synthesis of Section~3.6 and are not statistically independent, since several can respond
to the same prior, base-rate, or label-shift structure.

\subsubsection{Primary transfer performance. }
The first layer evaluates source-trained models in the target contexts, reporting AUROC, average
precision, and the Brier score so that ranking and probability-scale behavior are examined together. It
characterizes practical transfer behavior but cannot, on its own, identify the mechanism of any observed
change.

\subsubsection{Oracle-gap diagnostic. }
The second layer compares a source-trained model with a target-trained oracle on the same held-out
target fold. The oracle gap, the performance difference between the two on the identical fold, probes
target-learnable multivariate relationship change through achievable performance rather than any single
association. The target context is split 80/20 with stratification (oracle training and test sizes of
234{,}445 and 58{,}612 in the temporal setting and 26{,}243 and 6{,}561 in the cross-segment setting,
against source training sizes of 134{,}804 and 80{,}634), so the shared test fold removes test-set
variation but not training-size differences: the gap reflects the training context together with sample
size and model capacity. Its sensitivity bounds are discussed in the Limitations.

\subsubsection{Calibration diagnostic. }
The third layer distinguishes ranking behavior from probability-scale mismatch, using ECE and the
calibration intercept and slope; ECE is computed at several equal-frequency bin counts, with the ten-bin
value as the locked reference. It also applies a diagnostic recalibration: a single intercept shift on
the logit scale chosen so that the mean predicted probability matches the observed target-context default
rate. Because this uses target-context label information, it is an oracle-style counterfactual rather
than a deployable procedure, and it serves only to separate a probability-scale or prevalence explanation
from deeper relationship change.

\subsubsection{Feature-label stability diagnostic. }
The fourth layer measures shifts in marginal feature-label associations relative to random-split
noise-baseline variation, interpretable per feature. It is marginal and complementary: it describes
univariate association changes but cannot establish full stability of $P(y \mid x)$. A Spearman measure
is reported only as a descriptive sensitivity check, not the locked band metric (univariate AUC strength).

\subsubsection{Positive-control sensitivity. }
The fifth layer injects known relationship shifts at increasing levels (weak, medium, strong: 5, 15, and
30 percent of target rows relabeled by a deterministic rule on the two perturbation features, which
changed 1.5, 4.6, and 9.2 percent of labels) and tests whether the diagnostic stack responds,
characterizing the sensitivity boundary of the protocol rather than validating it. The perturbation uses \texttt{dti\_n} and \texttt{fico\_n}, two interpretable
numeric covariates, as a transparent design choice rather than the strongest predictors; the observed
default label is preserved and the synthetic labels held separately, so the feature matrix and real
outcome remain intact.

\subsection{Rule-based multi-signal synthesis}

The outputs of the five layers are interpreted together using a pre-specified set of decision
rules, as a multi-signal synthesis rather than as independent confirmation. The complete rule set,
labeled A through G, was specified and committed before any official experiment and is condensed
in Table~\ref{tab:rules}: the antecedents and readings follow the committed rule text, whereas the
governance-use phrases after each semicolon are the conceptual mapping of Section~5.3 and were not
part of the locked rules. The rules map the combined pattern of primary performance, oracle gap,
feature-label stability, and positive-control response onto a pre-specified reading; they are
diagnostic descriptors rather than mutually exclusive verdicts, so where more than one applies, all
applicable readings are reported. No rule or threshold was added or edited after the official outputs
were seen; the only later change to the locked documents, committed after the primary-performance
stage and before the diagnostic stages, named the intermediate bands between the locked cutoffs
without altering them.

Rules A through D describe the real-data pattern over ranking stability, the oracle gap, calibration
behavior after recalibration, and feature-label stability. Rules E and F are evidential gates set by the
positive control, since a null real-data finding is credible only if the positive control shows the
diagnostic is sensitive. Rule G is the bounded-evidence reading for a stable pattern with a working
positive control. The locked thresholds that operationalize these terms are: AUROC is stable when the in-domain minus
transfer drop is below 0.02 (borderline from 0.02 to 0.05, degraded at 0.05 or more); an oracle gap is
small below 0.02, moderate/borderline from 0.02 to 0.05, and large at 0.05 or more; feature-label
instability is low at no more than twice its noise baseline, moderate/borderline between two and three
times, and high at three times or more; calibration is poor when ECE10 exceeds 0.05, the absolute
intercept exceeds 0.5, or the slope lies outside $[0.8, 1.25]$; recalibration fixes most of it when
ECE10 falls by at least 50 percent to at most 0.05; and the positive control is detected at a level when
its oracle gap exceeds the real temporal gap by at least 0.05 with a monotone weak-to-strong increase.
These were pre-specified as heuristic diagnostic anchors rather than calibrated to portfolio cost, and we
report them as such. A post-hoc sensitivity
analysis examined nearby AUROC, oracle-gap, and calibration cutoffs while holding the remaining locked
antecedents fixed; its outcome is reported in Section~4.6.

\begin{table*}[!tb]
\centering
\caption{The locked decision-rule set (A through G), pre-specified before any official experiment.
In the third column, the text after the semicolon is conceptual governance guidance added for this paper.}
\label{tab:rules}
\resizebox{\textwidth}{!}{%
\begin{tabular}{p{0.8cm} p{8.2cm} p{7.4cm}}
\hline
\textbf{Rule} & \textbf{Locked antecedent condition} & \textbf{Pre-specified reading; conceptual governance use} \\
\hline
A & AUROC stable, oracle gap small, and feature-label instability low &
No evidence of strong $P(y\,|\,x)$ supervision drift in this setting; routine monitoring \\
B & Raw calibration poor but intercept/base-rate recalibration fixes most of it &
Main deployment risk is prior/base-rate shift affecting calibration, not ranking failure;
validate a deployable recalibration before any retraining \\
C & Oracle gap large or feature-label instability high while AUROC remains stable &
Localized relationship drift may exist but ranking is preserved by redundant predictors;
recalibrate and investigate the relationship change \\
D & Calibration remains poor after recalibration and oracle gap is large &
Both prior/base-rate shift and deeper $P(y\,|\,x)$ drift may be present; retrain or escalate \\
E & Positive control not detected at medium or strong injection &
Diagnostic framework is insufficient; do not make strong drift-detection claims \\
F & Positive control detected only at strong injection, not at weak or medium &
Diagnostic detects large shifts but may miss subtle drift; real-data null findings interpreted
cautiously \\
G & All real-data diagnostics stable and positive control works &
Limited evidence of harmful supervision drift under the locked tests; the contribution is the
protocol and the bounded evidence, not a failure claim \\
\hline
\end{tabular}%
}
\end{table*}

\subsection{Artifact availability}

The locked protocol, decision rules, and run manifests are documented and implemented in code, covering
the five diagnostic layers, the rule-based synthesis, and the positive control, so that reported values
can be reproduced and the protocol applied unchanged to
new proxy-labeled settings. The dataset is the public LendingClub release of
\textcite{arizagarzon2024lendingclub}. The locked specifications were committed to version control before the first official run, so the
order of specification and official runs is documented in the repository history. Every
stage writes a manifest recording the command, the commit at execution, package versions, and the
SHA-256 hashes of its input files; the pre-run data-verification manifest records the SHA-256
fingerprint of the raw analysis file, which recomputes from the file used here. The protocol
implementation, locked settings, run manifests, and result tables are publicly available at
{\def\UrlBreaks{\do\/\do\-\do\.}\url{https://github.com/mehrdad-shoeibi/supervision-drift-credit}} and
archived on Zenodo \parencite{shoeibi2026artifact}.

\section{Results}

\subsection{Primary transfer performance}

Primary ranking performance is stable across the transfer settings. Mean AUROC ran from 0.612 to 0.659
(Table~\ref{tab:performance}), and no setting showed a positive AUROC degradation of 0.02 or more
relative to the in-domain reference: every model-level in-domain minus transfer difference was at or below zero,
ranging from $-0.0169$ to $-0.0006$. Temporal AUROC was slightly higher than the in-domain reference for
all three models; we do not read that as improvement or as absence of distribution shift. Bootstrap 95
percent intervals for each AUROC (2{,}000 resamples of the test fold, recorded with the repository
results) have half-widths of 0.0082 to 0.0092 for the in-domain and cross-segment rows and 0.0022 to
0.0023 for the temporal rows; the other diagnostics are reported as point values per model and seed,
with the noise baselines of Section~4.4 as the only sampling reference for the stability metrics.

Probability-scale behavior differs from ranking. Mean Brier scores were worse in the temporal setting
than in-domain by about 0.047 to 0.049, while the cross-segment evaluation, whose observed default rate
is lower rather than higher, shows slightly better Brier scores. The higher temporal Brier is consistent
with the prevalence and probability-scale pattern examined by the calibration diagnostic below, and is
not treated here as independent evidence of relationship drift.

\begin{table*}[!tb]
\centering
\caption{Primary transfer performance across locked evaluation settings. In-domain rows are the
reference; difference columns are not applicable (n/a) for those rows.}
\label{tab:performance}
\begin{tabular}{l l r r r r r}
\hline
\textbf{Context} & \textbf{Model} & \textbf{AUROC} & \textbf{Avg. prec.} & \textbf{Brier} & \textbf{$\Delta$AUROC} & \textbf{$\Delta$Brier} \\
\hline
in\_domain & hgb & 0.6524 & 0.2406 & 0.1268 & n/a & n/a \\
in\_domain & logreg & 0.6365 & 0.2326 & 0.1278 & n/a & n/a \\
in\_domain & rf & 0.6121 & 0.2093 & 0.1304 & n/a & n/a \\
temporal & hgb & 0.6592 & 0.3552 & 0.1751 & $-0.0068$ & $0.0484$ \\
temporal & logreg & 0.6534 & 0.3499 & 0.1747 & $-0.0169$ & $0.0469$ \\
temporal & rf & 0.6222 & 0.3157 & 0.1798 & $-0.0101$ & $0.0494$ \\
cross\_segment & hgb & 0.6530 & 0.2104 & 0.1115 & $-0.0006$ & $-0.0153$ \\
cross\_segment & logreg & 0.6408 & 0.2036 & 0.1123 & $-0.0043$ & $-0.0156$ \\
cross\_segment & rf & 0.6133 & 0.1829 & 0.1145 & $-0.0012$ & $-0.0159$ \\
\hline
\end{tabular}
\\[2pt]
{\footnotesize $\Delta$AUROC is in-domain minus transfer; $\Delta$Brier is transfer minus in-domain.}
\end{table*}

\subsection{Oracle-gap diagnostic}

The temporal oracle gap in AUROC ranged from 0.008343 to 0.016399 (mean 0.013340); the cross-segment
gap from $-0.007268$ to 0.006090 (mean $-0.001827$). No gap reached the large threshold of 0.05 and all
temporal gaps were below the small threshold of 0.02, so the settings show no large oracle advantage,
arguing against large detectable relationship drift. The small or negative cross-segment gaps are not
read as proof of no drift, since that oracle is trained on the smaller target segment.

The locked oracle-gap rule is defined on AUROC. Average precision and the Brier score, computed on the
same folds as supporting descriptive metrics, are likewise small: temporal means of 0.014653 (average
precision) and 0.007610 (Brier, source minus oracle, so positive favors the oracle), and cross-segment
means of $-0.005187$ and 0.000291, which differ in sign but agree that there is no material oracle
advantage. The supporting gaps do not suggest a qualitatively different picture from the AUROC-based
oracle-gap analysis.

\subsection{Calibration diagnostic}

\begin{figure}[!tb]
    \centering
    \includegraphics[width=\linewidth]{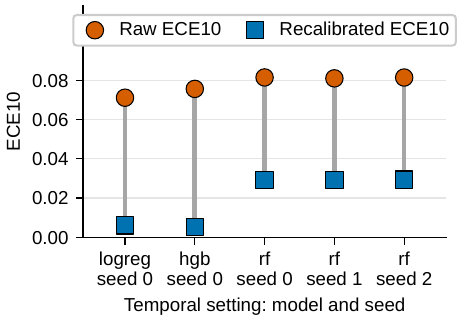}
    \caption{Temporal calibration diagnostic: raw versus diagnostic-recalibrated ECE10 for each model
    and seed. The recalibration is oracle-style and not deployable.}
    \label{fig:calibration}
\end{figure}

In the temporal setting, raw ECE10 ranged from 0.071088 to 0.081410; after recalibration it ranged from
0.005148 to 0.029311, a reduction of roughly 64 to 93 percent. Under the locked thresholds, all temporal
rows were poorly calibrated and all met the criterion that recalibration fixes most of the error. This
recalibration is an oracle-style counterfactual using target-context label information, not a deployable
correction. Figure~\ref{fig:calibration} shows the paired raw and recalibrated temporal ECE10 values.

The temporal random forest models showed calibration slopes outside the locked acceptable range (0.6362
to 0.6393; their intercepts, $-0.0457$ to $-0.0404$, were within range), and low slopes also appear in
the in-domain reference (slope 0.5797 to 0.5848, intercept $-0.6209$ to $-0.6107$), so the slope pattern
is a model-specific artifact rather than drift evidence. Overall, the temporal
calibration evidence is consistent with a prevalence and probability-scale mismatch that is largely
reducible in this diagnostic setting. Section~5.4 states why this release cannot identify what generated
that mismatch.

\subsection{Feature-label stability}

Under the locked numeric metric, the temporal ratio was 2.4895 (moderate/borderline) and the
cross-segment ratio 1.7518 (low); under the categorical metric both were low (temporal 1.4343,
cross-segment 0.8578). A descriptive Spearman-based check, not the locked band metric, gave a higher
temporal ratio of 4.8307 and is a caveat only.

\subsection{Positive-control sensitivity}

\begin{figure}[!tb]
    \centering
    \includegraphics[width=\linewidth]{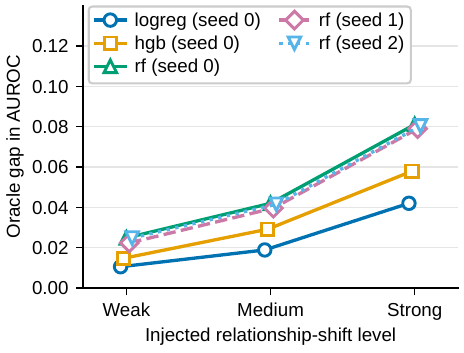}
    \caption{Positive-control sensitivity: oracle gap in AUROC across injected shift levels for each
    model and seed. Series are offset horizontally for legibility. Descriptive sensitivity curve, not a
    binary detection rule.}
    \label{fig:poscontrol}
\end{figure}

The positive-control curve injects a known relationship shift into synthetic 2016 labels at three locked
levels (weak, medium, strong) and measures the oracle gap. For every model and seed, the gap increased
monotonically across levels (Figure~\ref{fig:poscontrol}). Under a descriptive same-model, same-seed
reading (strong-level gap versus the real temporal gap plus 0.05), the random-forest curves exceeded the
threshold across all seeds, logistic regression did not, and histogram gradient boosting was close but
below; per-seed values appear in Figure~\ref{fig:poscontrol}. Under an aggregate-mean reading, the strong level exceeded the mean real temporal gap by
0.0547, while medium and weak did not (0.0209, 0.0061); this aggregate treats the three random-forest seeds
as separate rows, and with equal weight per model family the strong-level margin is 0.0484, just below
0.05, so the aggregate reading depends on the aggregation unit. The injection levels, model and seed plan, metric, comparator,
margin, and monotonicity requirement were locked before the positive control was run; the aggregation
unit, the row alignment, and the decision to report a curve rather than a binary flag were fixed
afterwards. We therefore report these as descriptive summaries rather than as a locked decision rule.

\subsection{Multi-signal synthesis}

Taken together, the locked real-data diagnostics do not show strong evidence of large detectable
relationship drift under the specified tests. Reading the layers jointly: ranking is stable, the
oracle-gap layer finds no large oracle advantage, the calibration layer is poorly calibrated raw but
largely reduced by intercept-only diagnostic recalibration, feature-label instability is mostly low to
moderate, and the positive control responds clearly only at strong injection. The closest reading is a
Rule B component in the temporal setting, stability consistent with Rule-A-style behavior in the
cross-segment setting, and an overall Rule-F-like sensitivity boundary, since the positive control was
detected clearly only at strong injection and only for the random-forest models, with no binary
positive-control decision introduced. Rule B classifies the observed diagnostic
signature; it does not establish the mechanism generating the prevalence difference.

A post-hoc sensitivity analysis over nearby cutoffs (AUROC stability at 0.01, 0.02, and 0.03; oracle-gap
bands of $(0.01, 0.03)$, $(0.02, 0.05)$, and $(0.03, 0.07)$; calibration ECE10 at 0.03, 0.05, and 0.08,
holding all other locked antecedents fixed) leaves this qualitative reading unchanged: every AUROC
difference remains stable, no context reaches a large oracle gap under any band, and Rules C and D are
not triggered. Two band labels move without changing a rule: under the tightest oracle-gap cutoff four
of five temporal rows move from small to intermediate, and at the loosest calibration cutoff one
temporal model moves from poorly to acceptably calibrated.

\section{Discussion}

\subsection{Principal finding and what the protocol adds}

The locked diagnostics do not show strong evidence of large detectable relationship drift under the
specified transfer settings. This is not a claim that drift is absent or that the conditional label
distribution is stable, but the multi-signal pattern of Section~4.6. A single metric is insufficient
here, since a stable AUROC can coexist with a calibration shift and the Brier score conflates
relationship change with prevalence and scale differences. The contribution is the layered protocol with
the interpretation discipline of the pre-specified rules; the positive control is a descriptive
sensitivity curve, not a formal power analysis, and the protocol is an auditing framework, not a
universally validated detector.

\subsection{Interpretation of the LendingClub evidence pattern}

Several readings require caution and are not treated as drift evidence on their own, as noted where each
is reported: the worse temporal Brier coincides with a higher later-vintage default rate, the slightly
higher temporal AUROC is not read as improvement, the cross-segment oracle is trained on the smaller
target segment, and the random-forest calibration parameters are a model-specific artifact.

The layers are complementary but not independent, which conditions how much their agreement adds. The
calibration layer, the Brier component of primary performance, and any prevalence-driven movement in the
oracle gap can all respond to the same prior shift, so agreement among them corroborates less than
agreement among mutually independent tests would. The oracle-gap and feature-label layers have
different failure modes: the former is limited by achievable target performance, whereas the latter
captures marginal associations only, so their agreement is more informative. We read convergence across layers
that share a failure mode as weaker evidence than convergence across layers that do not.

A further boundary applies to the temporal reading. The protocol can characterize an observable
diagnostic signature, but attributing that signature to a specific data-generating mechanism requires
information not contained in the analyzed release. Diagnosis and attribution are therefore separate
questions; the present artifact addresses the former but does not resolve the latter (Section~5.4).

\subsection{Implications for credit-risk decision support and model-risk governance}

For model-risk governance of proxy-labeled credit-risk models, apparent stability in a single headline
metric is incomplete evidence, and a multi-signal design with pre-specified rules reduces post-hoc
storytelling.

To make the intended use concrete, we trace how a model-risk analyst would read the outputs of
Section~4. From that pattern the analyst concludes that discrimination is intact, that the dominant
observed change is a prevalence and probability-scale mismatch (Rule B) whose cause is not established,
and that the evidence is bounded because the sensitivity floor leaves subtler drift unexcluded. Under the conceptual action mapping,
this pattern would motivate evaluating a deployable recalibration, distinct from the oracle-style one
used here for diagnosis, before considering retraining or escalation, together with continued monitoring
and targeted label collection. This mapping is conceptual guidance following from the meaning of each
layer, not a decision rule this paper validates; the appropriate action in any real portfolio depends on
cost, regulatory, and operational factors outside our scope. It also clarifies why an explicitly
non-deployable recalibration is still decision-relevant: by testing whether the observed error is
consistent with a prevalence explanation, it tells the team whether a deployable recalibration is worth
building before committing to the larger cost of retraining.

At a broader societal level, supervision drift can matter even when aggregate discrimination appears
stable, because probability-scale errors may affect pricing, approval thresholds, and other downstream
decisions. If such errors vary across borrower populations or with changing portfolio composition, they
may contribute to unequal access or terms without any visible collapse in headline AUROC. The present
study does not conduct a subgroup-fairness analysis, so these implications motivate monitoring rather
than constitute an empirical fairness claim.

\subsection{Limitations}

Several boundaries condition the interpretation of this study, and we state each once here. The most
important is empirical scope: the evaluation uses a single public LendingClub dataset and no second
external dataset. This preserves analytical depth but is a genuine limitation, so the absence of
large-drift evidence here should not be generalized to other credit-risk or lending settings.

A specific threat conditions the temporal reading. The dataset documentation states that loans in
transitory states were removed, so the release retains only loans whose status was resolved by the end of
the source's considered time window \autocite{arizagarzon2024lendingclub}; that window's endpoint is not
documented. The analyzed export contains no contractual term, no loan-resolution status, and no outcome
snapshot date, so we cannot determine whether later vintages are differentially selected by outcome
maturity, nor decompose the observed proxy-label prevalence difference (0.1560 in 2013 versus 0.2328 in
2016) into genuine population change, maturity-related selection, other cohort-composition effects, or a
mixture. The locked design documentation identified censoring and loan-maturity effects as potential
sources of bias across vintages before the official runs. A post-hoc descriptive check of within-vintage origination-month
patterns was inconclusive: directions varied across vintages and cannot identify a maturity mechanism. We therefore report the temporal diagnostic signature as
observed while treating its mechanism as unidentified in this release.

Three additional methodological boundaries constrain what the layers can support. The feature-label stability diagnostic
is marginal and univariate, so it constrains but cannot certify the conditional label distribution
$P(y\,|\,x)$. The diagnostic recalibration uses target-context label information and is therefore an
oracle-style counterfactual for analysis, not a deployable correction. The positive control is a
descriptive sensitivity curve whose response is model- and level-dependent, not a binary detector or a
formal power analysis; its aggregation and row-alignment choices were not locked as a decision rule.

The oracle-gap layer is itself bounded by model capacity, feature information, label noise, sample size,
and the locked thresholds, so a small gap limits the evidence for large drift but does not prove absence
of conditional shift. This floor is tightened by low base discrimination: with AUROC near
0.65 the audited relationship is weak, and subtle drift in a weak relationship is intrinsically harder to
detect, a condition the positive control is designed to quantify rather than hide. The cross-segment
oracle is also trained on the smaller target segment; since a smaller training set should understate the
oracle's achievable performance, the observed cross-segment oracle gap may be understated, which weakens
the ``no large gap'' reading rather than strengthening it.

Finally, the protocol is conditioned on the chosen transfer contexts, models, metrics, thresholds, and
decision rules; different choices could yield different sensitivity boundaries. Taken together, these
limitations mean smaller or subtler relationship drift cannot be ruled out: the diagnostics constrain
what would have been visible but do not certify that the conditional relationship is unchanged.

\subsection{Future work}

Future work should apply the same locked protocol to additional datasets and domains, ideally with
pre-registered dataset selection so that replication does not become a search for favorable outcomes.
Natural extensions include a release carrying loan term and resolution status, which would permit a
maturity-matched temporal comparison, richer model classes, and validating the diagnostic-to-governance
mapping.

\section{Conclusion}

This paper contributes a locked, multi-signal audit protocol for supervision drift in proxy-labeled
credit-risk prediction, framed as a design-science artifact and demonstrated on one public LendingClub
dataset across temporal and cross-segment transfer. The protocol yields bounded evidence: the overall
pattern is closest to Rule F, in that positive-control sensitivity appeared only at strong injection and
only for some models, with a Rule B-style probability-scale and prevalence component in the
temporal setting and stability consistent with Rule-A-style behavior in the cross-segment setting, while
Rule D is not triggered under the locked thresholds. The protocol characterizes that signature; it does not attribute it to a mechanism,
and the analyzed release cannot separate population change from maturity-related selection or other
cohort effects. We treat this as an informative audit outcome rather than a failure, since the artifact
reports both what it detects and where its sensitivity is limited. We do not conclude that relationship
drift is absent; smaller or subtler drift remains possible within the sensitivity boundary.

\printbibliography

\end{document}

%% file: hicss-packages.tex
\usepackage[letterpaper]{geometry}
\usepackage{hicss}
\usepackage{times}
\usepackage[none]{hyphenat}
\usepackage{url}
\usepackage{latexsym}
\usepackage{indentfirst}
\usepackage{graphicx}
\graphicspath{{images/}}
\usepackage[
    style=apa,
  ]{biblatex}